\documentclass[10pt,letterpaper]{article}

\usepackage[top=0.85in,left=2.2in,footskip=0.75in,marginparwidth=2in]{geometry}
\usepackage[utf8]{inputenc}
\usepackage{cite}
\usepackage{nameref,hyperref}
\usepackage[right]{lineno}
\usepackage{microtype}
\DisableLigatures[f]{encoding = *, family = * }

\raggedright
\usepackage{changepage}
\usepackage[aboveskip=1pt,labelfont=bf,labelsep=period,singlelinecheck=off]{caption}
\makeatletter
\renewcommand{\@biblabel}[1]{\quad#1.}
\makeatother

\usepackage{lastpage,fancyhdr,graphicx}
\usepackage{epstopdf}
\fancyheadoffset[L]{2.25in}
\fancyfootoffset[L]{2.25in}

\usepackage{color}
\definecolor{Gray}{gray}{.25}
\usepackage{graphicx}
\usepackage{sidecap}
\usepackage{wrapfig}
\usepackage[pscoord]{eso-pic}
\usepackage[fulladjust]{marginnote}
\reversemarginpar
\usepackage[numbers]{natbib}

\usepackage{graphicx}
\usepackage{multirow}
\usepackage{algorithm}
\usepackage{algorithmic}
\usepackage{amssymb}
\usepackage{amsmath}
\usepackage{epsfig}
\usepackage{epstopdf}
\usepackage{centernot}
\usepackage{amsfonts}
\usepackage{color}
\usepackage{soul}
\usepackage{adjustbox}
\usepackage{subfig}
\usepackage{url}
\usepackage{mathrsfs}

\usepackage{mathtools}
\DeclarePairedDelimiter{\ceil}{\lceil}{\rceil}

\usepackage{afterpage}

\usepackage{array}
\usepackage{booktabs}

\begin{document}

% Full title of the paper (Capitalized)

\begin{flushleft}
	{\Large
		\textbf\newline{Center Emphasized Visual Saliency and a Contrast-based Full Reference Image Quality Index}
	}
	\newline

Md Abu Layek\textsuperscript{1},
Sanjida Afroz\textsuperscript{2},
TaeChoong Chung\textsuperscript{1},
Eui-Nam Huh\textsuperscript{1}

\bigskip
\bf{1} Department of Computer Science and Engineering, Kyung Hee University Global Campus, Yongin-17104, Republic of Korea;
\\
\bf{2} Independent Researcher, MS in Psychology, University of Dhaka, Bangladesh
\\
\bigskip
* abulayek@yahoo.com, layek@khu.ac.kr, sanjida.psy@gmail.com

\end{flushleft}

% Abstract (Do not insert blank lines, i.e. \\) 
\section*{abstract}
Objective image quality assessment (IQA) is imperative in the current multimedia-intensive world, in order to assess the visual quality of an image at close to a human level of ability. Many~parameters such as color intensity, structure, sharpness, contrast, presence of an object, etc., draw human attention to an image. Psychological vision research suggests that human vision is biased to the center area of an image and display screen. As a result, if the center part contains any visually salient information, it draws human attention even more and any distortion in that part will be better perceived than other parts. To the best of our knowledge, previous IQA methods have not considered this fact. In this paper, we propose a full reference image quality assessment (FR-IQA) approach using visual saliency and contrast; however, we give extra attention to the center by increasing the sensitivity of the similarity maps in that region. We evaluated our method on three large-scale popular benchmark databases used by most of the current IQA researchers (TID2008, CSIQ~and LIVE), having a total of 3345 distorted images with 28~different kinds of distortions. Our~method is compared with 13 state-of-the-art approaches. This comparison reveals the stronger correlation of our method with human-evaluated values. The prediction-of-quality score is consistent for distortion specific as well as distortion independent cases. Moreover, faster processing makes it applicable to any real-time application. The MATLAB code is publicly available to test the algorithm and can be found online \protect\footnote{http://layek.khu.ac.kr/CEQI}.

%\linenumbers

\section*{Introduction}
\label{intro}
Computer-based automatic image quality assessment (IQA) has been sought after for decades because numerous image and video applications need this assessment to automate their quality maintenance. To date, IQA research has been significantly advanced, however, this is still an active area of research to bring the methods closer to human-level ability. In the literature, there are three principle IQA approaches. No-reference image quality assessment (NR-IQA) uses a single distorted image without any reference image, whereas in reduced-reference IQA (RR-IQA), partial information of a reference image is given. The third category is full-reference IQA (FR-IQA) where the complete reference image is given along with the distorted one. In this paper, we deal with FR-IQA.

Table \ref{Table:ComparingIQAs} presents a brief survey on several state-of-the-art IQA approaches, which we compared in this paper. We made a separate column for the pooling strategy because recent IQA research tends to combine multiple features where pooling plays an important role. Early pixel-based, faster IQA methods such as mean squared error (MSE) and peak signal-to-noise ratio (PSNR) consider neither the human visual system (HVS) nor any aspects of human perception. Thus, those approaches fail to achieve good correlation with human assessment \cite{girod1991psychovisual, wang2009mean}. Two images with the same PSNR or MSE may be perceived in totally different ways by a human observer. However, humans are the ultimate receiver of images; as a result, the search for methods that can achieve a closer correlation with humans is ongoing. Wang et al., in their revolutionary work on the structural similarity index or SSIM \cite{wang2004image}, argued that human visual perception is highly sensitive to structural information. The SSIM index incorporates luminance, contrast, and structural comparison information and achieves a very good correlation with the mean opinion scores (MOS) of human observers. Inspired by the success of SSIM, several extended versions, such as the multi-scale structural similarity for image quality assessment (MS-SSIM)~\cite{wang2003multiscale} and Information content weighting for perceptual image quality assessment (IW-SSIM) \cite{wang2011information}, were proposed by the same research group. IW-SSIM utilizes an image pyramid to decompose the original and distorted images into versions of varying scales, and then computes the information content from the images. Finally, it finds the quality score using the information content as a weighting function.

Based on shared information between the reference and distorted images, Sheikh et al. proposed the information fidelity criteria (IFC) \cite{sheikh2005information}  and the visual information fidelity (VIF) \cite{sheikh2005visual}. The most apparent distortion (MAD) approach \cite{larson2010most} separates images based on the distortion and applies either a detection-based strategy or an appearance-based strategy. Some of the methods, such as the noise quality measure (NQM) \cite{damera2000image} and the visual signal-to-noise ratio (VSNR) \cite{chandler2007vsnr}, take into account the HVS by incorporating interactions among different visual signals. In contrast, other approaches, including the popular feature similarity index or FSIM \cite{zhang2011fsim}, emphasize phase congruency \cite{kovesi1999image,liu2007phase,saha2013perceptual}. FSIM uses the image gradient as a secondary feature and local quality maps are weighted by phase congruency to obtain the final score. The image gradient has been used effectively in a number of other works \cite{chen2006gradient,zhu2012image}. Xue et al., in their gradient magnitude standard deviation (GMSD)~\cite{xue2014gradient}, used the gradient magnitude with a different pooling strategy, by applying the standard deviation, and Alaei et al. adopted a similar approach for assessing document images \cite{alaei2015document}. Both examples prove the effectiveness of standard deviation pooling, however, the authors of the GMSD approach showed that standard deviation (SD) pooling is not effective for all types of methods. Wang et al. proposed the multi-scale contrast similarity deviation (MCSD) metod \cite{wang2016multiscale}, which can be termed as a continuation of SSIM and MS-SSIM, since it also uses the root mean square (RMS) contrast similarity; however, they employed standard deviation pooling for the final score.   

Meanwhile, inspired by vision-related psychological research, visual saliency (VS)-based IQA methods \cite{zhang2014vsi,ma2008saliency}, which utilize different kinds of visual saliencies \cite{hou2007saliency,duan2011visual,zhang2012sr}, have attracted researchers' attention. In the visual saliency index (VSI) method \cite{duan2011visual}, VS is used as both a quality map and the weighting function at the pooling stage. The spectral residual similarity index (SR-SIM) \cite{zhang2012sr} uses the spectral residual saliency, which makes the approach very fast while maintaining a competitive correlation with the mean opinion score. Combining VS with other features has also become popular \cite{li2013color,jia2018contrast}. Li  et al. proposed an approach that combines VS and FSIM  while, recently,  Jia  et al. used contrast and spectral residual saliency as well as summation-based SD pooling.

% start a new page
\newpage
% change it to landscape
\paperwidth=\pdfpageheight
\paperheight=\pdfpagewidth
\pdfpageheight=\paperheight
\pdfpagewidth=\paperwidth
\newgeometry{layoutwidth=297mm,layoutheight=210 mm, left=2.7cm,right=2.7cm,top=1.8cm,bottom=1.5cm, includehead,includefoot}
\fancyheadoffset[LO,RE]{0cm}
\fancyheadoffset[RO,LE]{0cm}
%%%%%%%%%%%%%%%%%%%%%%%%%%%%%%%%%%%%%%%%%%%%%%

\begin{table}[H]
	\centering \small
	\caption{Overview of the compared image quality assessment (IQA) methods.}
	\label{Table:ComparingIQAs}
	\vspace{-6pt}
	\scalebox{0.9}[0.9]{
		\begin{tabular}{>{\centering}m{2.5cm}>{\centering}m{5.5cm}m{5cm}<{\centering}m{9cm}<{\raggedright}}
			\toprule
			\textbf{IQA} \textbf{Method}   & \textbf{Principle } \textbf{Consideration}			& \textbf{Pooling} \textbf{Used} &  \multicolumn{1}{c}{\textbf{Comments}}\\
			\midrule
			PSNR    	 	& Pixel-by-pixel error      	& Average     & Primitive method, does not consider HVS, poor correlation with humans, low computation, widely used.\\ \midrule
			\cite{wang2004image}  SSIM    		& Luminance, contrast, structure & Average & Milestone method to consider structural information that better represents HVS.     \\ \midrule
			\cite{wang2003multiscale} MS-SSIM    		& Multi-scale structure     	& Weighted sum & An extension to SSIM   that incorporates variations of viewing conditions and capable of multi-scale assessment.    \\ \midrule
			\cite{wang2011information} IW-SSIM    	 	& Information content extraction     & Weighted by information content  & Main emphasis is on information content extraction which is applicable to other methods as well, uses image~pyramid.   \\\midrule
			\cite{sheikh2005visual} MAD     			& Most visible distortion     	& Weighted product & A novel strategy consisting of two phases to detect the most apparent distortion.    \\ \midrule
			\cite{zhang2011fsim} FSIM    		& Phase congruency, Gradient magnitude & Weighted average & A state-of-the art IQA approach using phase congruency and the gradient magnitude weighted by phase congruency to calculate the final score.    \\\midrule
			\cite{xue2014gradient} GMSD    		& Image gradient     			& Standard deviation & Very fast assessment approach after PSNR showing competitive performance with other state-of-the-art~approaches.   \\ \midrule
			\cite{duan2011visual} VSI    			& Visual saliency (VS)   			& Weighted average & Introduced visual saliency for IQA where VS is used for both the quality map and weighting function at the pooling~stage.   \\ \midrule
			\cite{wang2016multiscale} MCSD    		& Multi-scale contrast     		& Standard deviation & Another fast method next to GMSD, but providing better~results.    \\ \midrule
			\cite{sheikh2005visual} VIF    			& Visual information extraction     & Average & Quantifies the extracted reference information from a distorted~image.    \\ \midrule
			\cite{damera2000image} NQM   			& Distortion and noise    			& Squared sum  & Considers HVS using distortion and noise, better than~PSNR.   \\ \midrule
			\cite{zhang2012sr} SR-SIM    		& Saliency and gradient     		& Weighted average & Uses spectral residual saliency and image gradient, which shows competitive performance.   \\ \midrule
			\cite{jia2018contrast} VSP    			& Saliency and contrast    		& Weighted sum of deviations  & Uses deviation pooling in a combined method.  \\ \midrule 
			CEQI (Proposed) & Saliency, contrast, center emphasis     & Weighted sum of deviations  & Gives special attention to the center area. \\ 
			\bottomrule		
		\end{tabular}}
		\\ 
		\begin{tabular}{@{}c@{}} 
			\multicolumn{1}{p{\linewidth-0.75cm}}{\footnotesize 
				PSNR: peak signal-to-noise ratio;
				SSIM: structural similarity index; 
				MS-SSIM: multi-scale structural similarity;
				IW-SSIM: information content weighted structural similarity;
				MAD:~most apparent distortion;
				FSIM: feature similarity index;
				GMSD: gradient magnitude standard deviation;
				VSI : visual saliency index;
				MCSD: multi-scale contrast similarity deviation ;
				VIF:~visual information fidelity;
				NQM: noise quality measure;
				SR-SIM : spectral residual similarity index ;
				VSP: visual saliency plus;						
				CEQI : center-emphasized quality index;
				HVS: human visua~system.
			}
		\end{tabular}
	\end{table}
	
	%%%%%%%%%%%%%%%%%%%%%%%%%%%%%%%%%%%%%%%%%%%%%%%
	% change everything back
	\newpage
	\restoregeometry
	\paperwidth=\pdfpageheight
	\paperheight=\pdfpagewidth
	\pdfpageheight=\paperheight
	\pdfpagewidth=\paperwidth
	\headwidth=\textwidth

\begin{figure}[H]	
	\centering
	\captionsetup[subfloat]{justification=centering, labelformat=empty}
	\begin{tabular}{ccc}
		
		\begin{tabular}{@{}c@{}}
			\subfloat['Sailing.bmp' from LIVE database \label{fig:WholeImage}]{
				\includegraphics[width=.36\linewidth]{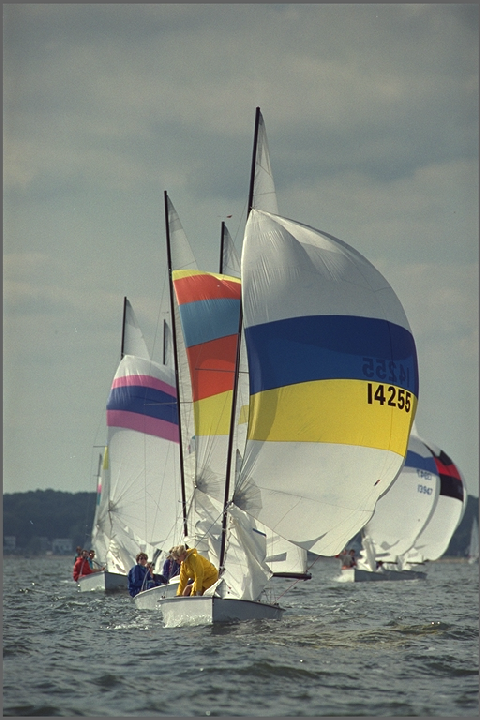}}
		\end{tabular}
		&
		\begin{tabular}{@{}c@{}}
			
			$\Longrightarrow		$
		\end{tabular}
		&	
		\begin{tabular}{@{}c@{}}
			\subfloat[Block01\label{fig:Block1}]{
				\includegraphics[width=.14\linewidth]{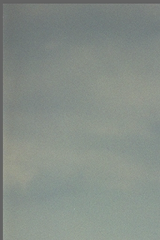}} 
			\subfloat[Block02\label{fig:Block2}]{
				\includegraphics[width=.14\linewidth]{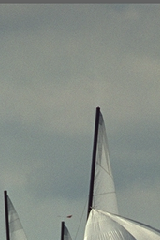}}
			\subfloat[Block03\label{fig:Block3}]{
				\includegraphics[width=.14\linewidth]{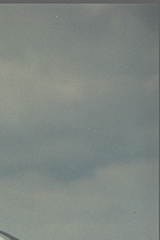}} \\
			\subfloat[Block04\label{fig:Block4}]{
				\includegraphics[width=.14\linewidth]{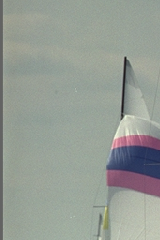}} 
			\subfloat[Block05\label{fig:Block5}]{
				\includegraphics[width=.14\linewidth]{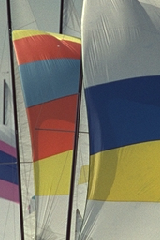}}
			\subfloat[Block06\label{fig:Block6}]{
				\includegraphics[width=.14\linewidth]{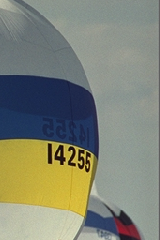}} \\
			\subfloat[Block07\label{fig:Block7}]{
				\includegraphics[width=.14\linewidth]{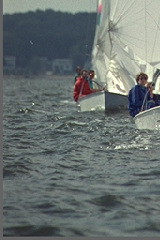}} 
			\subfloat[Block08\label{fig:Block8}]{
				\includegraphics[width=.14\linewidth]{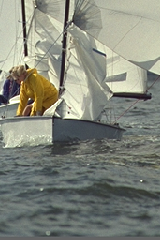}}
			\subfloat[Block09\label{fig:Block9}]{
				\includegraphics[width=.14\linewidth]{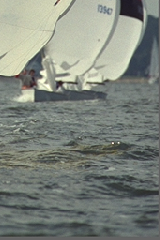}} 
		\end{tabular}
		
	\end{tabular}
	\caption{The image 'Sailing.bmp' is split into nine blocks, where Block05 is the center area.}
	\label{fig:center}
\end{figure}

In the context of the HVS, center bias in early eye movements is an established fact in psychological vision research \cite{langford1936people, mannan1997fixation, parkhurst2002modeling, tatler2007central}. Bindemann found that eye movement is biased not only to the scene center but also to the screen center \cite{bindemann2010scene}. As a result, if a scene appears at the center of the screen, it will receive the most attention. For example, in Figure \ref{fig:center}, the human eye will first move to the Block05 region and if that part has visually important information then it will attract even more attention. As a result, people will be more sensitive to the distortions in this region. To the best of our knowledge, there is no research in IQA considering this center bias for quality assessment. 

In this paper, we propose a new method for IQA which accounts for the center emphasis in HVS. In the proposed method, we first obtain both the contrast and VS similarity maps for the entire image. To give center emphasis, we find the VS similarity map of the mid-region and apply element-wise multiplication in the mid-part to raise the similarity deviation there. However, for the contrast similarity, we apply element-wise squaring in the center part. Contrast is a local quality map, so we do not calculate the contrast of the mid-area separately. On the other hand, VS is a global quality map, and thus it is calculated differently in the mid-region. The final score is obtained by performing weighted summation of the standard deviations on both of the similarity maps; further details with mathematical equations are given in section \ref{sec:ceqi}.

\vspace{-6pt}

We evaluated our proposed method on three popular benchmark databases for IQA research and compared it with 13 other state-of-the-art approaches. The results in terms of the correlations with human evaluated scores show that the method proposed by us outperforms the other approaches, with a reasonable amount of processing time.

This paper is organized as follows. Section \ref{sec:bg} describes some underlying theories and related techniques. Section \ref{sec:ceqi} explains the proposed center-emphasized assessment approach, and the results with relevant discussions are presented in Section \ref{sec:res}. Finally, the paper is concluded in Section \ref{sec:con}.
%%%%%%%%%%%%%%%%%%%%%%%%%%%%%%%%%%%%%%%%%%%%%%%%%%%%%%%%%%%%%%%%%%%%%

\section{Background}
\label{sec:bg}

In this section, we briefly review the underlying theories on which the content of this paper relies, including the spectral residual visual saliency similarity, contrast similarity, standard deviation pooling, and relevant evaluation metrics.

\subsection{Spectral Residual Visual Saliency Similarity}
\label{sec:bg:srvss}
In the human visual system (HVS), some interesting or salient regions of an image receive more attention than other parts. Detection of image saliency itself is an active field in vision research. As such, the human is more sensitive to these salient parts and any distortion in these parts attracts more intense attention, which makes it an important feature in IQA methods. There are a lot of saliency detection techniques available \cite{cong2018review}, among which spectral residual saliency detection \cite{hou2007saliency} is a very fast approach. We~adopt the saliency map generator described in the SR-SIM \cite{zhang2012sr} and the visual saliency plus contrast (VSP) \cite{jia2018contrast} approaches. 

For an image $f(x,y)$, according to Reference~\cite{hou2007saliency}, the spectral residual saliency (SRS) is computed as~follows:  

\begin{equation}
\label{eq:SRS1}
M_f(u,v)=abs[\mathscr{F}\{f(x,y)\}(u,v)]
\end{equation}

\begin{equation}
\label{eq:SRS2}
A_f(u,v)=angle[\mathscr{F}\{f(x,y)\}(u,v)]
\end{equation}

\begin{equation}
\label{eq:SRS3}
L_M(u,v)=\log\{M_f(u,v)\}
\end{equation}

\begin{equation}
\label{eq:SRS4}
R_f(u,v)=L_M(u,v)-h_n(u,v)*L_M(u,v)
\end{equation}

\begin{equation}
\label{eq:SRS5}
SRS(x,y)=g(x,y)*[\mathscr{F}^{-1}\{\exp(R_f+jA_f)\}(x,y)]^2,
\end{equation}
where $\mathscr{F}$ and $\mathscr{F}^{-1}$ denote the Fourier and inverse Fourier transforms, respectively; $abs(.)$ and $angle(.)$ return the magnitude and argument of a complex number, respectively; $h_n(u,v)$ is an $n\times n$ mean filter; $g(x,y)$ is a Gaussian filter; and $*$ denotes the convolution operation.  

In this way, we calculate the SRS for both the reference and distorted images denoted by $SRS_r(x,y)$ and $SRS_d(x,y)$, respectively. Then, the spectral residual visual saliency similarity ($SRVSS(r,d)$) is calculated as:

\begin{equation}
\label{eq:SRVSS}
SRVSS(r,d)=\frac{2SRS_r(x,y)\odot SRS_d(x,y)+c_1}{SRS_r(x,y)^2+SRS_d(x,y)^2+c_1},
\end{equation}   
where $\odot$ is the element-wise multiplication, $^2$ is the element-wise squaring and $c_1$ is a positive constant used to increase the calculation stability.

\subsection{Contrast Similarity}
\label{sec:bg:cs}

Contrast is a basic perceptual attribute of an image \cite{peli1990contrast} which varies greatly over the image, and the contrast map (CM) contains the spatial distribution of those varying values. There are many ways devised for calculating CMs and in this paper, we adopted the RMS contrast from SSIM, since it achieves better performance for natural images. The RMS contrast map $C_X$ of an image signal $X$ is given by:

\begin{equation}
\label{eq:CM1}
C_X=\left[\frac{1}{N-1}\sum_{i=1}^{N}(x_i-\mu_X)^2\right]^{\frac{1}{2}},
\end{equation} 
where $N$ is the total number of pixels in the image, $x_i$ is the intensity of pixel $i$, and $\mu_X$ is the mean intensity, defined as:
\begin{equation}
\label{eq:CM3}
\mu_X = \frac{1}{N} \sum_{i=1}^{N}x_i.
\end{equation}  

Again, using Equation (\ref{eq:CM1}), we obtain the contrast maps $C_r$ and $C_d$ for both reference and distorted images, respectively, and find the contrast similarity $CS(r,d)$ as follows:

\begin{equation}
\label{eq:CS}
CS(r,d)=\frac{2C_r \odot C_d+c_2}{C_r^2+C_d^2+c_2},
\end{equation}

where similar to Equation (\ref{eq:SRVSS}), $\odot$ is the element-wise multiplication, $^2$ is the element-wise squaring and $c_2$ is a positive constant used to increase the calculation stability.

\subsection{Standard Deviation Pooling}
\label{sec:bg:sdp}

As discussed in the introduction, standard deviation (SD) pooling achieves very good performance in specific cases and is adopted by several successful methods. Jia et al.  conducted an experiment with several other combinations of pooling and found that SD pooling provides the best correlation. The final quality score ($QS$) is calculated using the following equation:
\begin{equation}
\label{eq:QS}
QS=\frac{1}{w_1+w_2}\Big[w_1\times std\{SRVSS(r,d)\}+w_2\times std\{CS(r,d)\}\Big],
\end{equation}

where $w_1$ and $w_2$ are weighting factors that specify the importance of VSS and CS, respectively. The standard deviations in the above equation are defined as:
\begin{equation}
\label{eq:QS2}
std(SRVSS(r,d))=\Big\{\frac{1}{M}\sum_{i=1}^{M}(SRVSS_i-\mu_{SRVSS})^2\Big\}^{\frac{1}{2}}
\end{equation}

\begin{equation}
\label{eq:QS3}
std(CS(r,d))=\Big\{\frac{1}{M}\sum_{i=1}^{M}(CS_i-\mu_{CS})^2\Big\}^{\frac{1}{2}},
\end{equation} 
where $M$ is the number of total elements in the similarity matrices;
${SRVSS_i}$ and ${CS_i}$ are the $i$th items; $\mu_{SRVSS}$ and $\mu_{CS}$ are the mean values of the ${SRVSS(r,d)}$ and ${CS(r,d)}$, respectively, and are given~by:

\begin{equation}
\label{eq:QS4}
\mu_{SRVSS} = \frac{1}{M} \sum_{i=1}^{M}SRVSS_i
\end{equation} 

\begin{equation}
\label{eq:QS5} 
\mu_{CS} = \frac{1}{M} \sum_{i=1}^{M}CS_i   
\end{equation}

\subsection{Evaluation Metrics}
\label{sec:bg:EM}

The performance of any IQA method is usually measured by the mean squared error and several other correlations with the subjective scores that are human evaluated values usually in the form of MOS, or their differential DMOS. However, to apply linear correlation, the two compared values should be on the same scale and perfectly linearly correlated \cite{ding2018general}. To ensure better fairness, before applying the linear correlation measurements, a logistic mapping function is used to convert the objective scores. We use the following nonlinear regression model as suggested by Sheikh \cite{sheikh2006statistical}.

\begin{equation}
\label{eq:DataFit}
q^,=\beta_1\Big\{\frac{1}{2}-\frac{1}{1+\exp(\beta_2(q-\beta_3))}\Big\}+\beta_4q+\beta_5,
\end{equation} 
where $q$ is the objective score calculated by a IQA method, $q^,$ is the  mapped value, and $\beta_i, i=1,2,3,4,5$ are the parameters that are tuned based on the relationship
between objective and subjective scores. We utilized the MATLAB function $nlinfit$ to find the optimal parameters. After the mapping is done, the subjective scores are then used with these mapped scores to find the correlation coefficients.

One of the widely adopted basic correlations is Pearson's linear correlation coefficient (PLCC) which is defined as follows:
\begin{equation}
\label{eq:PLCC}
PLCC(o,s)=\frac{\sum_{i=1}^m(o_i-\mu_o)(s_i-\mu_s)}{\big\{\sum_{i=1}^m(o_i-\mu_o)^2\big\}^\frac{1}{2} \big\{\sum_{i=1}^m(s_i-\mu_s)^2\big\}^\frac{1}{2}},
\end{equation} 
where $m$ is the number of distorted images; $o$ and $s$ are vectors of the objective and subjective scores, respectively; and $\mu_o$ and $\mu_s$ are the mean scores, defined by:

\begin{equation}
\label{eq:PLCCm1}
\mu_o = \frac{1}{m} \sum_{i=1}^{m}o_i
\end{equation}

\begin{equation}
\label{eq:PLCCm2}
\mu_s = \frac{1}{m} \sum_{i=1}^{m}s_i.
\end{equation} 

In our case, the objective scores $o$ are actually the mapped scores using Equation (\ref{eq:DataFit}). If the nonlinear mapping in Equation (\ref{eq:DataFit}) is to be avoided, then rank order coefficients can be used. The most popular Spearman's rank-order correlation coefficient (SROCC) is defined as:
\begin{equation}
\label{eq:SROCC}
SROCC(o,s)=PLCC(rank(o),rank(s)).
\end{equation}
The function $rank()$ of a vector returns a rank-vector, where the $i$-th entry contains the relative rank of the $i$-th item in the original vector. 

Another popularly adopted rank order coefficient is Kendall's rank-order correlation coefficient (KROCC), which is given as below:
\begin{equation}
\label{eq:KROCC}
KROCC(o,s)=\frac{C-D}{m(m-1)/2},
\end{equation} 
where $C$ is the number of concordant pairs that are consistently correlated between objective and subjective scores; and $D$ is the number of discordant pairs.

The root mean square error (RMSE) is also commonly adopted and is defined as:
\begin{equation}
\label{eq:RMSE}
RMSE(o,s)=\left\{\frac{1}{m}\sum_{i=1}^m(o_i-s_i)^2\right\}^\frac{1}{2}
\end{equation} 
A larger value for PLCC, SROCC, and KROCC indicates that the corresponding method is better. On~the other hand, a smaller value of the RMSE is a sign of a superior IQA.
%%%%%%%%%%%%%%%%%%%%%%%%%%%%%%%%%%%%%%%%%%%%%%%%%%%%%%%%%%%%%%%%%%%%%  

\section{Proposed Center-Emphasized Quality Assessment} 
\label{sec:ceqi} 

\begin{figure}[H]
	\centering
	\includegraphics[width=0.88\textwidth]{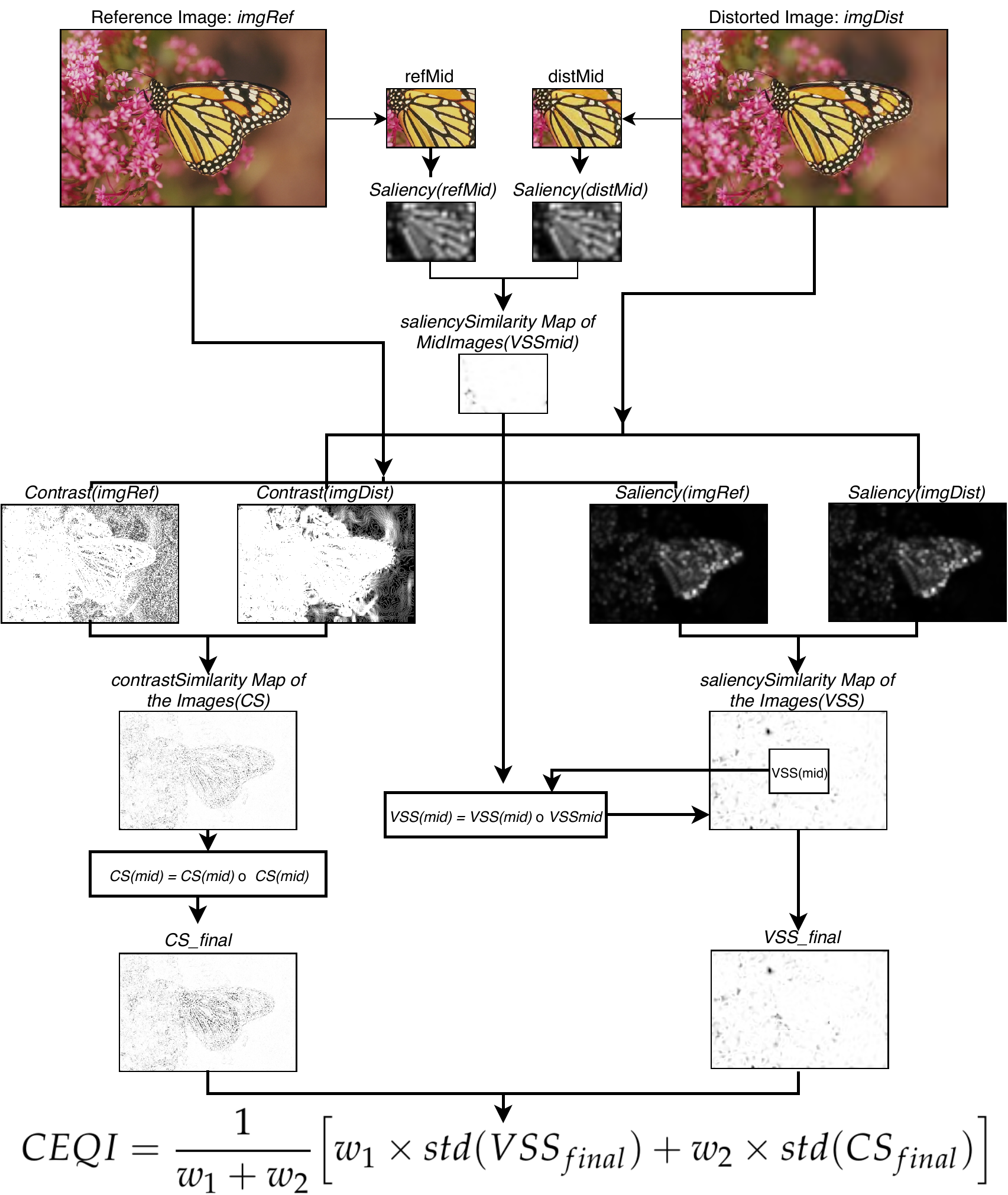}
	\caption{Flow diagram of the proposed center-emphasized approach. }
	\label{fig:flow_diagram}
\end{figure}

The general flow diagram of our proposed method is presented in Figure \ref{fig:flow_diagram}.
At first, the center parts of both the reference and distorted images are extracted. To do this, we split the image in $3\times 3$~image blocks as shown in Figure \ref{fig:center}, and the fifth block, which resides in the middle both horizontally and vertically, is taken as the center area. If the original image dimension is $(H \times W)$, then the corresponding dimension for the center block becomes $(H_{mid} \times W_{mid})$, where:
\begin{equation}
\label{eq:midI}
H_{mid}=\ceil[\bigg]{\frac{H}{3}} \text{ \;\;  and \;\;  } W_{mid}=\ceil[\bigg]{\frac{W}{3}}.
\end{equation} 

The center block is defined as a rectangular area identified by two corner points $(x_{min}, y_{min}) \text{\; and \;} (x_{max}, y_{max})$,
where:
\begin{equation}
\label{eq:rec}
%\begin{split}
x_{min}=\ceil[\bigg]{\frac{H}{3}},\;
y_{min}=\ceil[\bigg]{\frac{W}{3}},\; 
x_{max}=\ceil[\bigg]{\frac{H}{3}}+H_{mid},\text{\; and \;}
y_{max}=\ceil[\bigg]{\frac{W}{3}}+W_{mid}
%\end{split}
\end{equation} 

%\begin{equation}
%\label{eq:rec}
%\begin{split}
%x_{min}=\ceil[\bigg]{\frac{H}{3}}\\
%y_{min}=\ceil[\bigg]{\frac{W}{3}}\\
%x_{max}=\ceil[\bigg]{\frac{H}{3}}+H_{mid}\\
%\text{\; and \;} y_{max}=\ceil[\bigg]{\frac{W}{3}}+W_{mid}
%\end{split}
%\end{equation} 

First, the saliency similarity maps for the full images and middle images are found using Equations (\ref{eq:SRS1})--(\ref{eq:SRVSS}) and are denoted as $VSS$ and $VSS_{mid}$, respectively. Simultaneously, the contrast similarity map for the full-size, $CS$, is also obtained. As discussed before, we do not derive the CS map for middle images.

Then, we increase the sensitivity of the center area within both of the maps. Let $VSS(mid)$ and $CS(mid)$ be the center areas of $VSS$ and $CS$, respectively. The updated middle parts will be determined as follows:

\begin{equation}
\label{eq:VSS_U}
VSS(mid)=VSS(mid)\odot VSS_{mid}
\end{equation} 

\begin{equation}
\label{eq:CS_U}
CS(mid)=CS(mid)\odot CS(mid),
\end{equation}
where $\odot$ is the element-wise multiplication.

With the updated middle portion, we obtain the finalized maps $VSS_{final}$ and $CS_{final}$, and using Equation (\ref{eq:QS}), we calculate the final quality score of the proposed method $CEQI$ as: 

\begin{equation}
\label{eq:CEQI}
CEQI=\frac{1}{w_1+w_2}\Big[w_1 \times std(VSS_{final})+w_2 \times std(CS_{final})\Big]
\end{equation}

\section{Results and Analysis}
\label{sec:res}

Experiments were carried out on three popular benchmark databases for IQA research---TID2008~\cite{ponomarenko2009tid2008}, CSIQ~\cite{larson2010categorical} and LIVE~\cite{sheikh2005live}.
Our approach was compared with 13 other state-of-the-art approaches as listed in Table \ref{Table:ComparingIQAs}. Basic information about the databases is given in Table \ref{Table:DatasetDescription} and the distortion information is recorded in Table \ref{Table:NoiseWiseDatasetDescription}.

\begin{table}[H]
	\caption{Basic information about the databases used for the experiments.}
	\centering
	%\resizebox{\columnwidth}{!}
	{	
		\includegraphics[width=.93\textwidth]{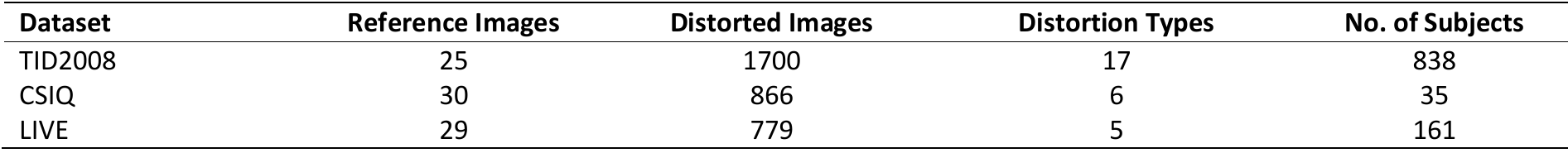}	
	}
	\label{Table:DatasetDescription}
\end{table}

\vspace{-12PT}

\begin{table}[H]
	\vspace{-6pt}
	\caption{Types of distortion used in each database.}
	\label{Table:NoiseWiseDatasetDescription}
	\centering \small
	
	\scalebox{0.8}[0.8]{
		\begin{tabular}{>{\centering}m{1cm}>{\centering}m{1cm}m{1cm}<{\centering}{|}m{5cm}<{}m{3cm}<{\centering}}
			\toprule
			\textbf{TID2008} & \textbf{CSIQ} & \textbf{LIVE} & \textbf{Type of distortion}                         & \textbf{Abbreviation} \\
			\toprule
			Y                & Y             & Y             & Additive Gaussian noise                             & AGN                   \\
			Y                & -             & -             & Additive noise in color components                  & ANC                   \\
			Y                & -             & -             & Spatially correlated noise                          & SCN                   \\
			Y                & -             & -             & Masked noise                                        & MN                    \\
			Y                & -             & -             & High frequency noise                                & HFN                   \\
			Y                & -             & -             & Impulse noise                                       & IN                    \\
			Y                & -             & -             & Quantization noise                                  & QN                    \\
			Y                & Y             & Y             & Gaussian blur                                       & GB                    \\
			Y                & -             & -             & Image denoising                                     & IDN                   \\
			Y                & Y             & Y             & JPEG compression                                    & JPEG                  \\
			Y                & Y             & Y             & JPEG2000 compression                                & JP2K                  \\
			Y                & -             & -             & JPEG transmission errors                            & JGTE                  \\
			Y                & -             & -             & JPEG2000 transmission errors                        & J2TE                  \\
			Y                & -             & -             & Non-eccentricity pattern noise                      & NEPN                  \\
			Y                & -             & -             & Local block-wise distortions of different intensity & LBD                   \\
			Y                & -             & -             & Mean shift (intensity shift)                        & MS                    \\
			Y                & Y             & -             & Contrast change                                     & CTC                   \\
			-                & -             & Y             & Fast fading Rayleigh                                & FF                    \\
			-                & Y             & -             & Additive pink Gaussian noise                        & AWPN \\
			\toprule                
		\end{tabular}}
	\end{table}
%%%%%%%%%%%%%%%%%%%%%%%%%%%%%%%%%%%%%%%%%%%%%%%%%%%%%%%%%%%%%%%%%%%%%%

For performance comparison, we use four commonly adopted metrics---Spearman’s rank-order correlation coefficient (SROCC), Kendall’s rank-order correlation coefficient (KROCC), Pearson's linear correlation coefficient (PLCC), and the root mean square error (RMSE)---which we defined in Section~\ref{sec:bg:EM}.

Table \ref{Table:compareTableOverall} compares the four metrics among the different IQA models, for all of the three databases. The top three values for each metric are typed in boldfaced and light-gray shaded; the top value is colored blue; the second highest value is colored red, and the third highest value is colored black. However, in the case of RMSE, coloring is done in a reverse way, {i.e.,} the lowest value is colored in blue and so on, since a lower RMSE implies a better method. We see that, for the biggest database, TID2008, our proposed method outperforms all other methods in all metrics. For the other two databases, it achieves competitive performance. We calculated the weighted averages of the SROCC, KROCC, PLCC, and RMSE using the number of distorted images to find the overall performance, as proposed in Reference~\cite{wang2011information}. It can be noticed that, compared to VSI and VSP, our approach shown better prediction accuracy with  (1.09\%,~0.3\%)-point, (2.44\%,~0.39\%)-point and (2.19\%,~0.22\%)-point higher overall SROC, KROC and PLCC values, respectively. The overall ranking based on performance is shown in Table \ref{Table:OverallRanking}.

\begin{table}[H]	
	\centering
	\caption{Performance comparison of IQA methods on three databases.}
	\label{Table:compareTableOverall}
	\vspace{-6pt}
	{	
		\includegraphics[width=.9\textwidth]{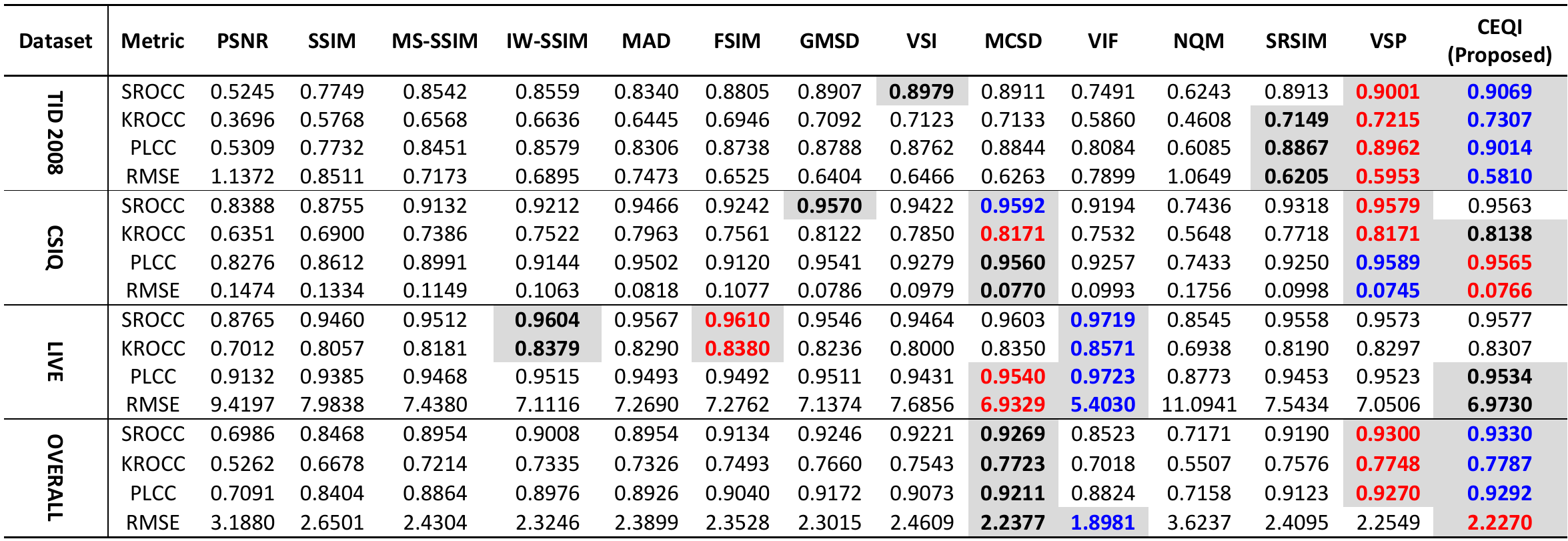}
	}	
	\begin{tabular}{@{}c@{}} 
		\multicolumn{1}{p{\linewidth-1.5cm}}{\footnotesize
			-- For each row, the first, second and third-ranked performances are highlighted respectively in blue, red and black colors.} \\
		\multicolumn{1}{p{\linewidth-1.5cm}}{\footnotesize
			-- For SROCC, KROCC and PLCC metrics, the higher the value, the better the method whereas for RMSE a lower score is better.}  
	\end{tabular}		
\end{table}

\vspace{-12PT}

\begin{table}[H]
	\caption{Overall performance ranking of the compared IQA methods.}
	\centering
	%\resizebox{\columnwidth}{!}
	{	
		\includegraphics[width=.80\textwidth]{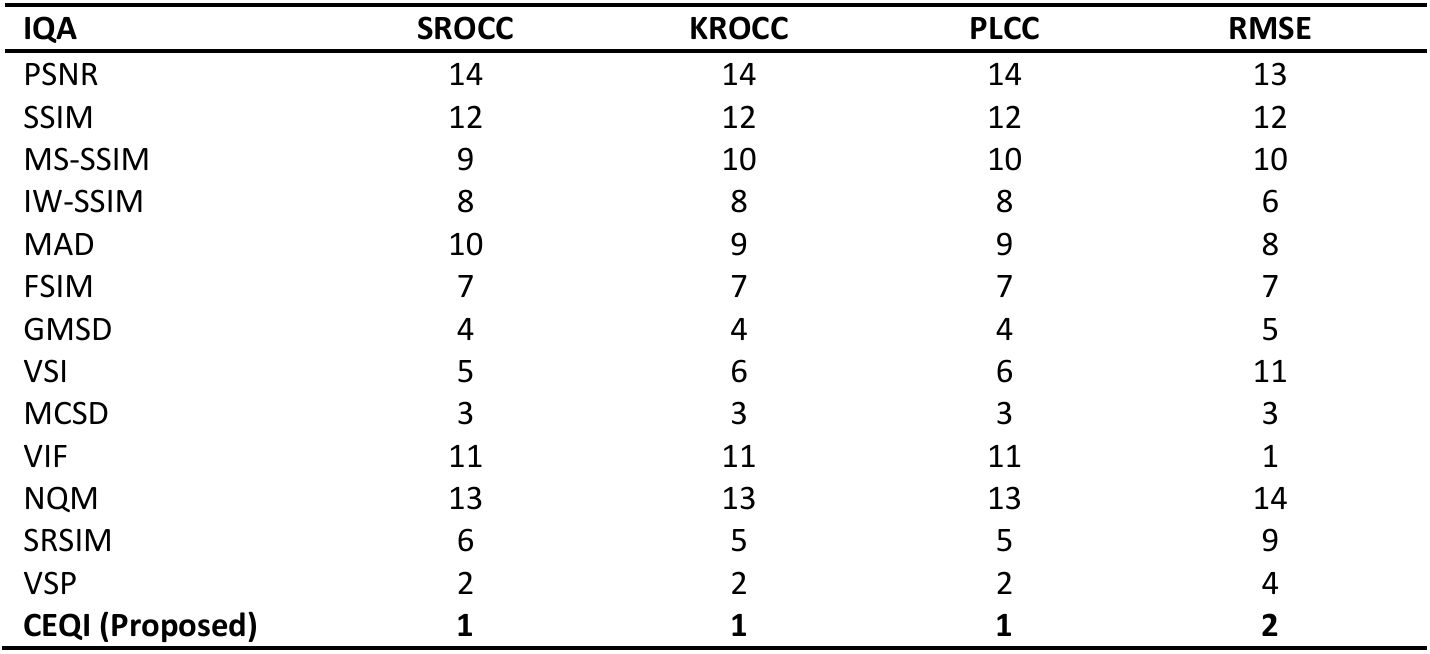}	
	}
	
	\footnotesize{-- Number 1 indicates the best performing method and 14 is the worst.}\\
	\label{Table:OverallRanking}	
\end{table}
%%%%%%%%%%%%%%%%%%%%%%%%%%%%%%%%%%%%%%%%%%%%%%%%%%%%%%%%%%%%%%%%%

%%%%%%%%%%%%%%%%%%%%%%%%%%%%%%%%%%%%%%%%%%%%%%%%%%%%%%%%%%%%%%%%%%%%%%%

Table \ref{Table:compareTableNoisewise} compares the SROCC performance for all distortion types; please refer to Table \ref{Table:NoiseWiseDatasetDescription} for a description of the abbreviations. We see that different methods perform better for different distortions and performance even varies between databases. This is the case because images are not affected equally by a specific type of distortion---it depends on the color, salient regions, and perhaps a combination of many other factors. Still, distortion-wise comparison gives us a good understanding of whether an IQA method is biased to some noise type or not. It can be seen that the proposed CEQI performs consistently well for all types of distortion; it is not too biased to any specific type of distortion, while retaining an average performance within the top 3 methods.

\begin{figure}[H]
	\centering
	\vspace{-20pt}
	%	\captionsetup[subfloat]{justification=centering, labelformat=empty}	
	\begin{tabular}{ccc}
		\subfloat[\label{fig:PridictSSIm}]{
			\includegraphics[width=.3\linewidth]{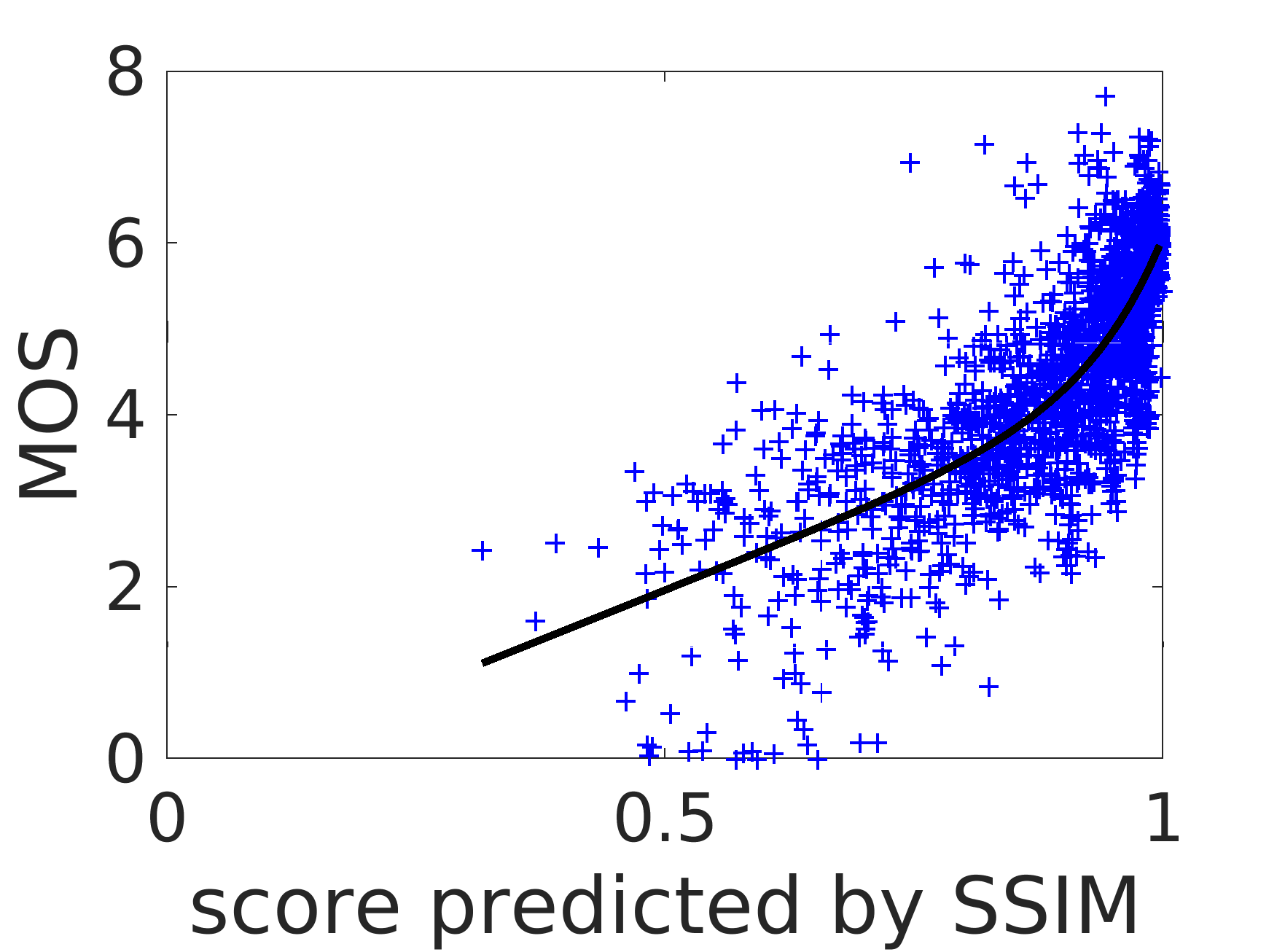}}
		&
		\subfloat[\label{fig:PridictMSSSIM}]{
			\includegraphics[width=.3\linewidth]{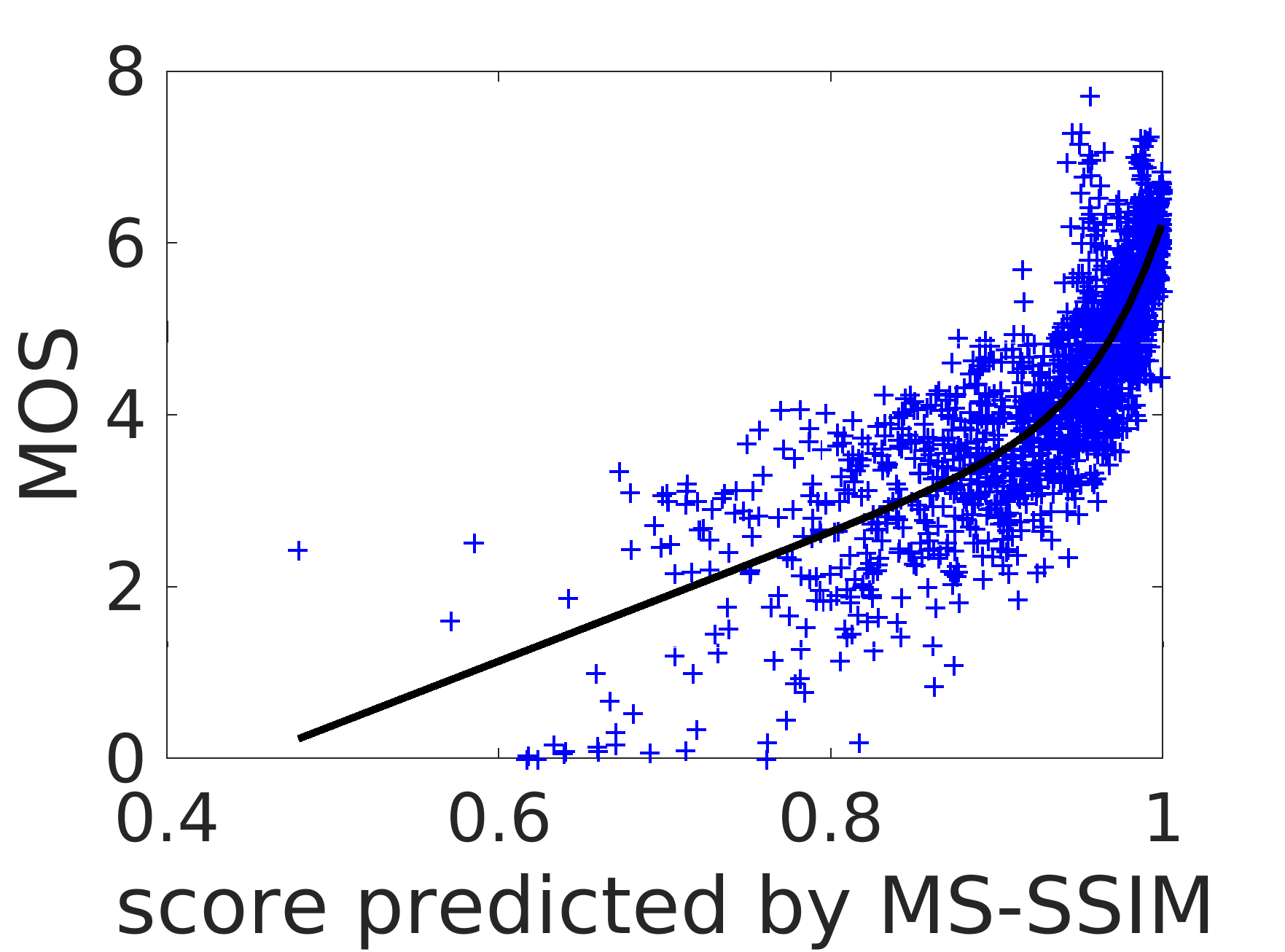}}
		&		\subfloat[\label{fig:PridictIWSSIM}]{
			\includegraphics[width=.3\linewidth]{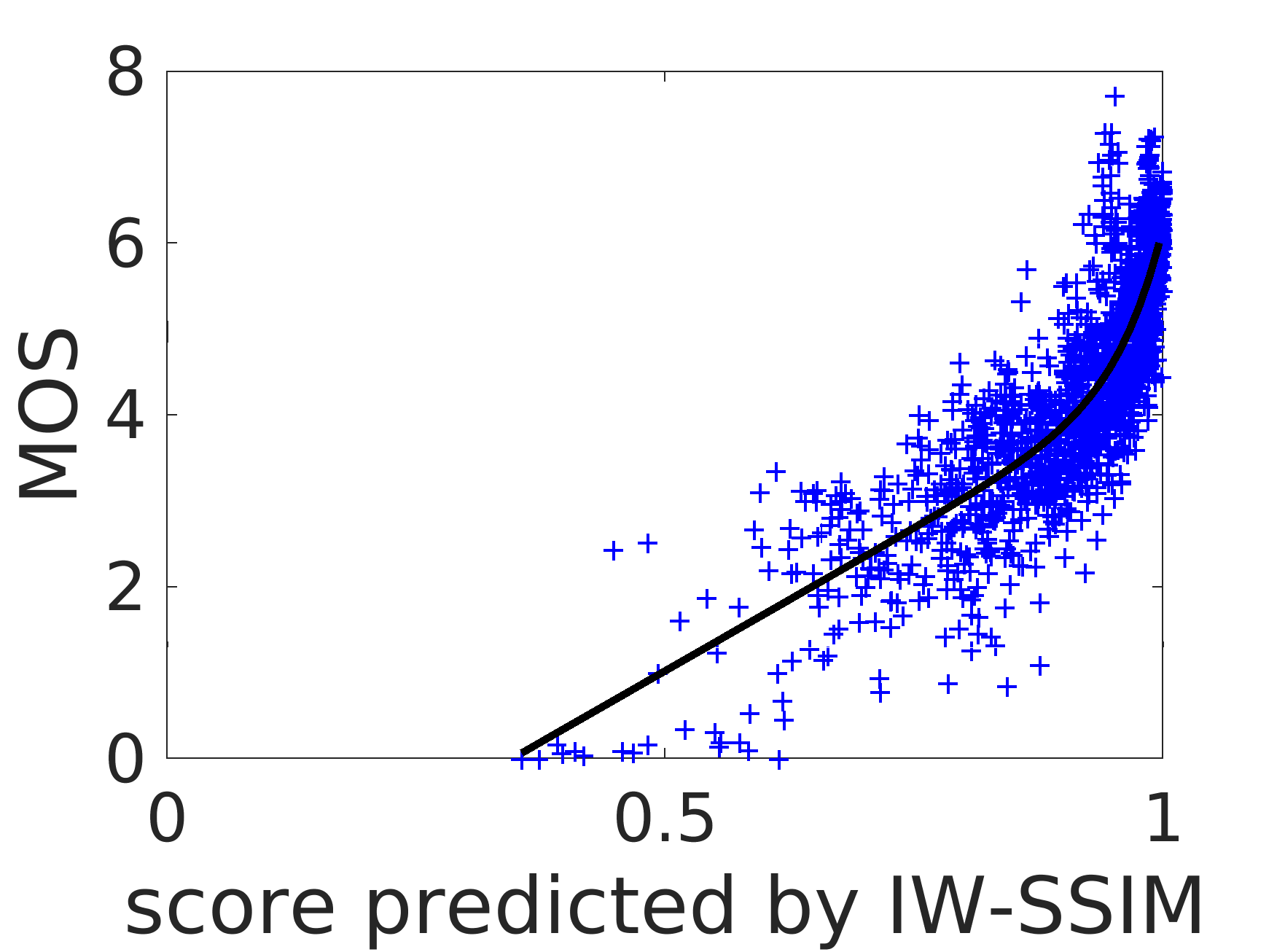}}\\

		\subfloat[\label{fig:PridictMAD}]{
			\includegraphics[width=.3\linewidth]{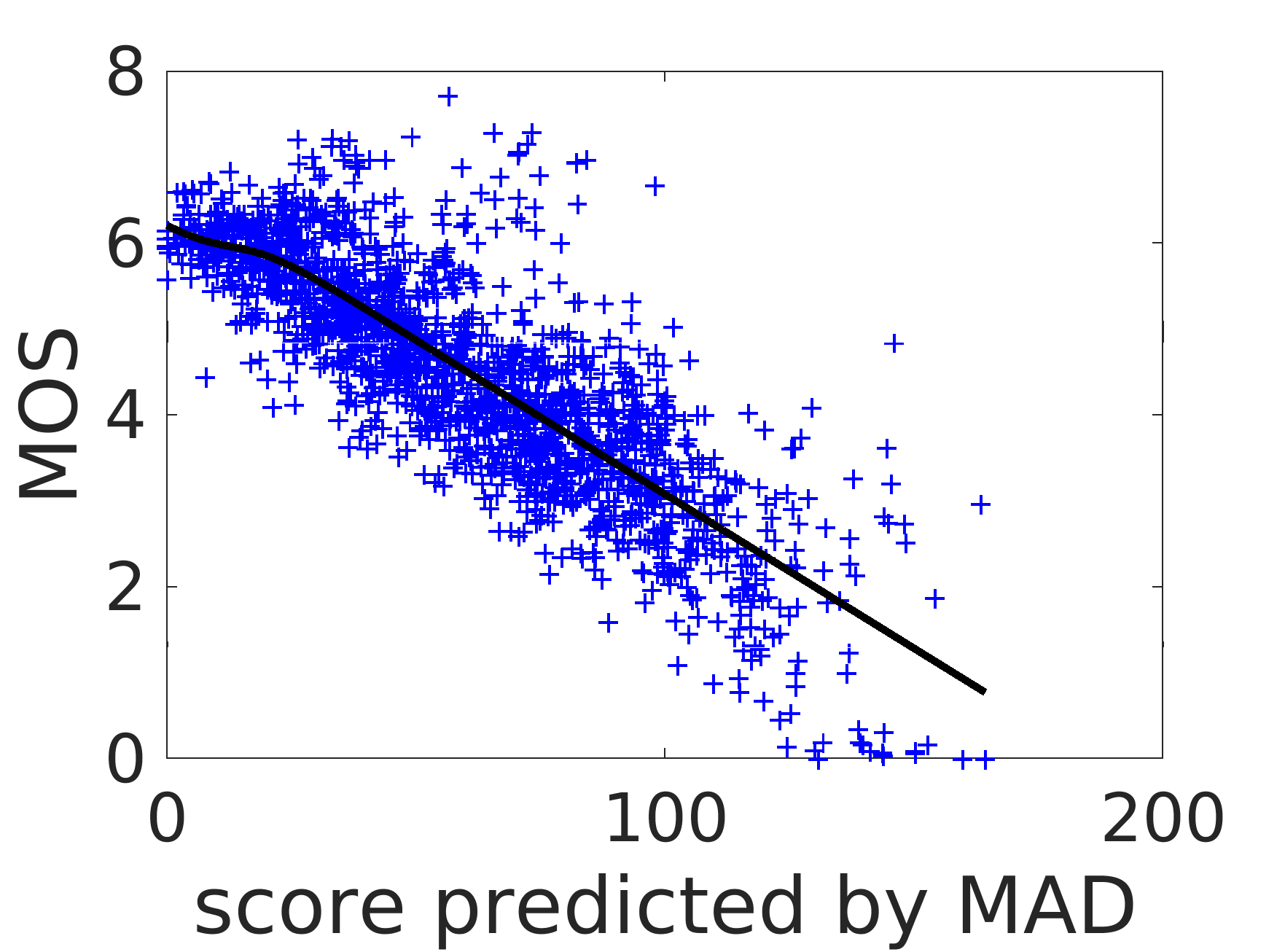}}
		&
		\subfloat[\label{fig:PridictFSIM}]{
			\includegraphics[width=.3\linewidth]{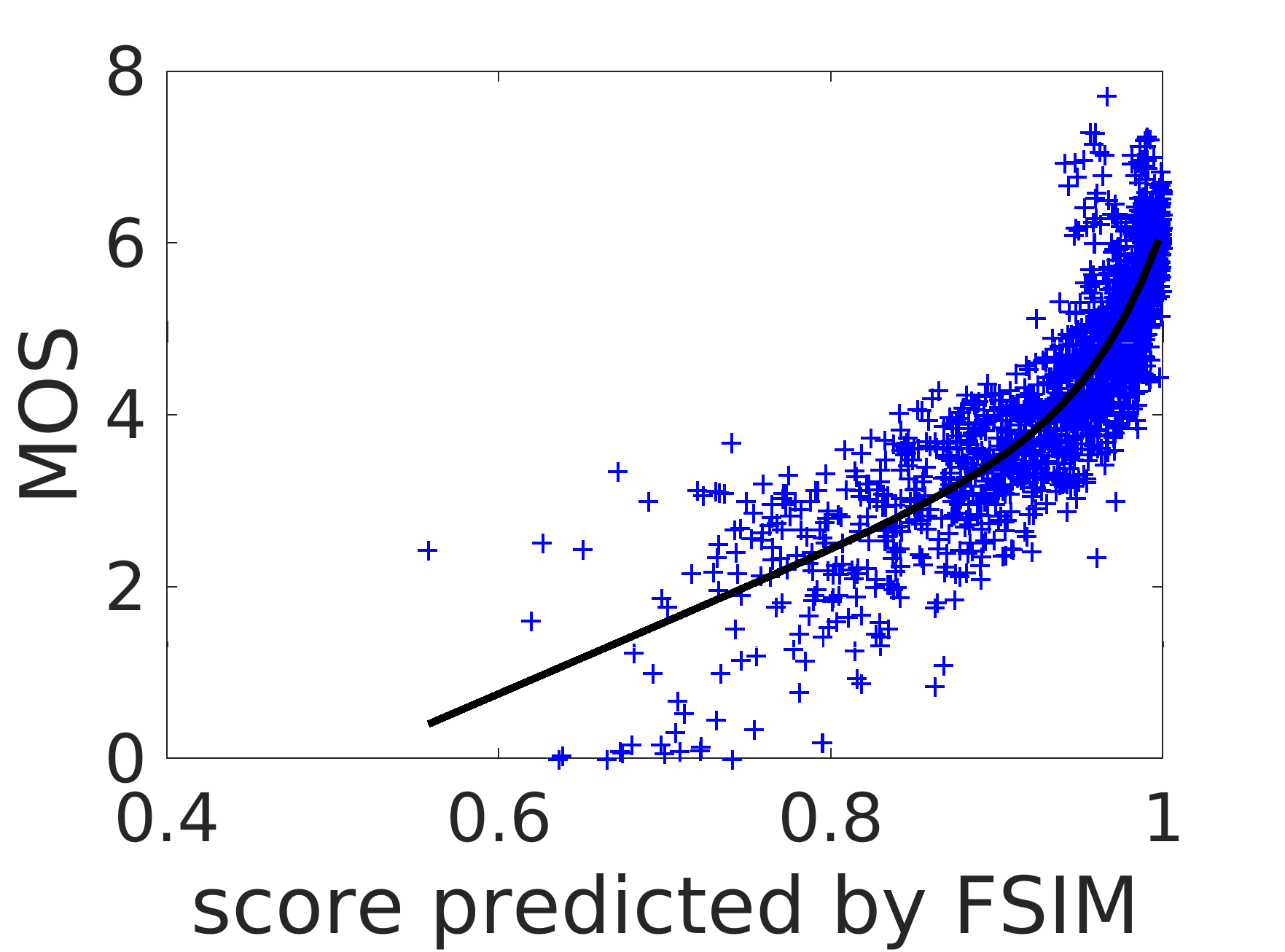}}
		&
		\subfloat[\label{fig:PridictGMSD}]{
			\includegraphics[width=.3\linewidth]{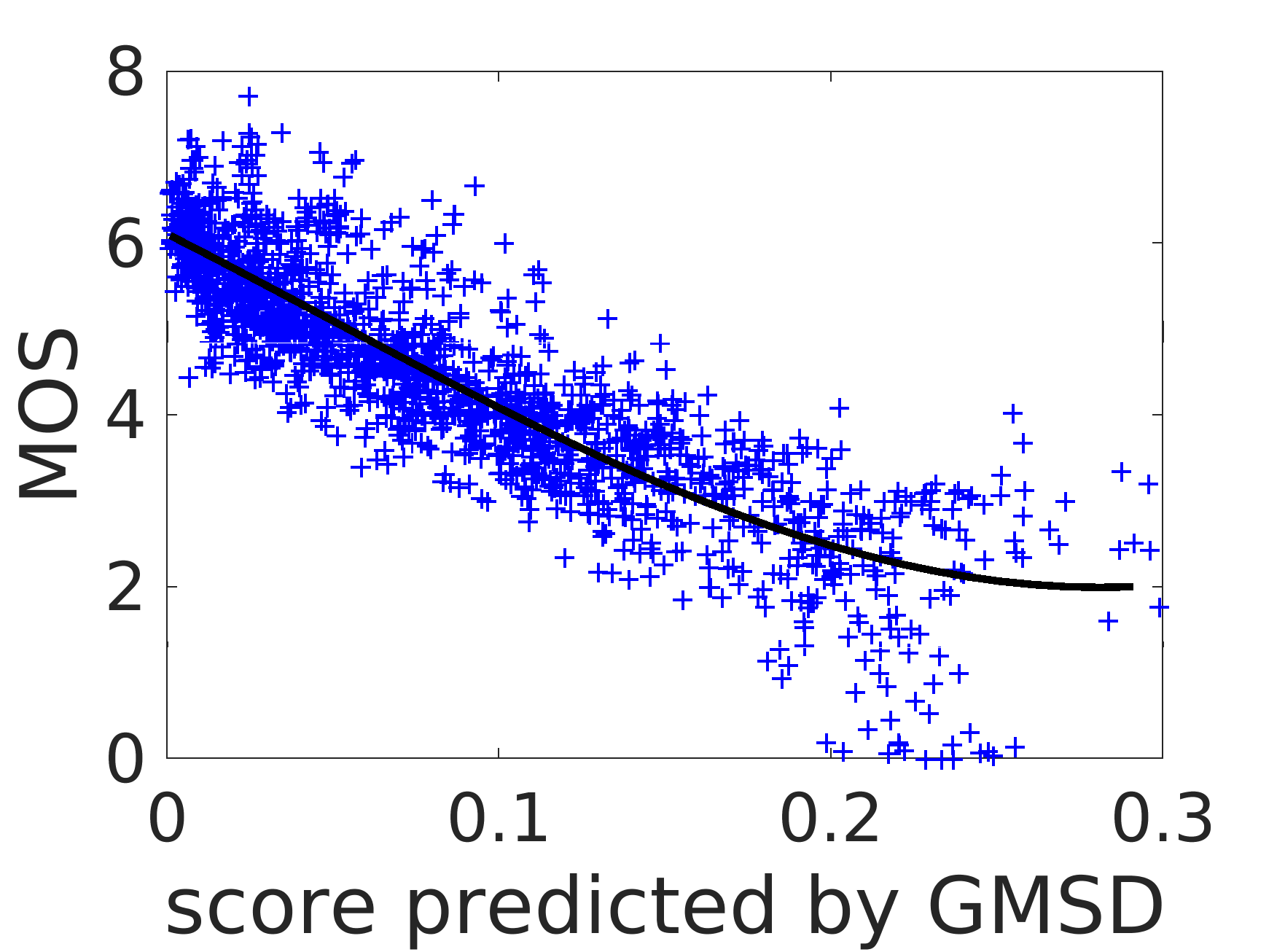}}\\

		\subfloat[\label{fig:PridictVSI}]{
			\includegraphics[width=.3\linewidth]{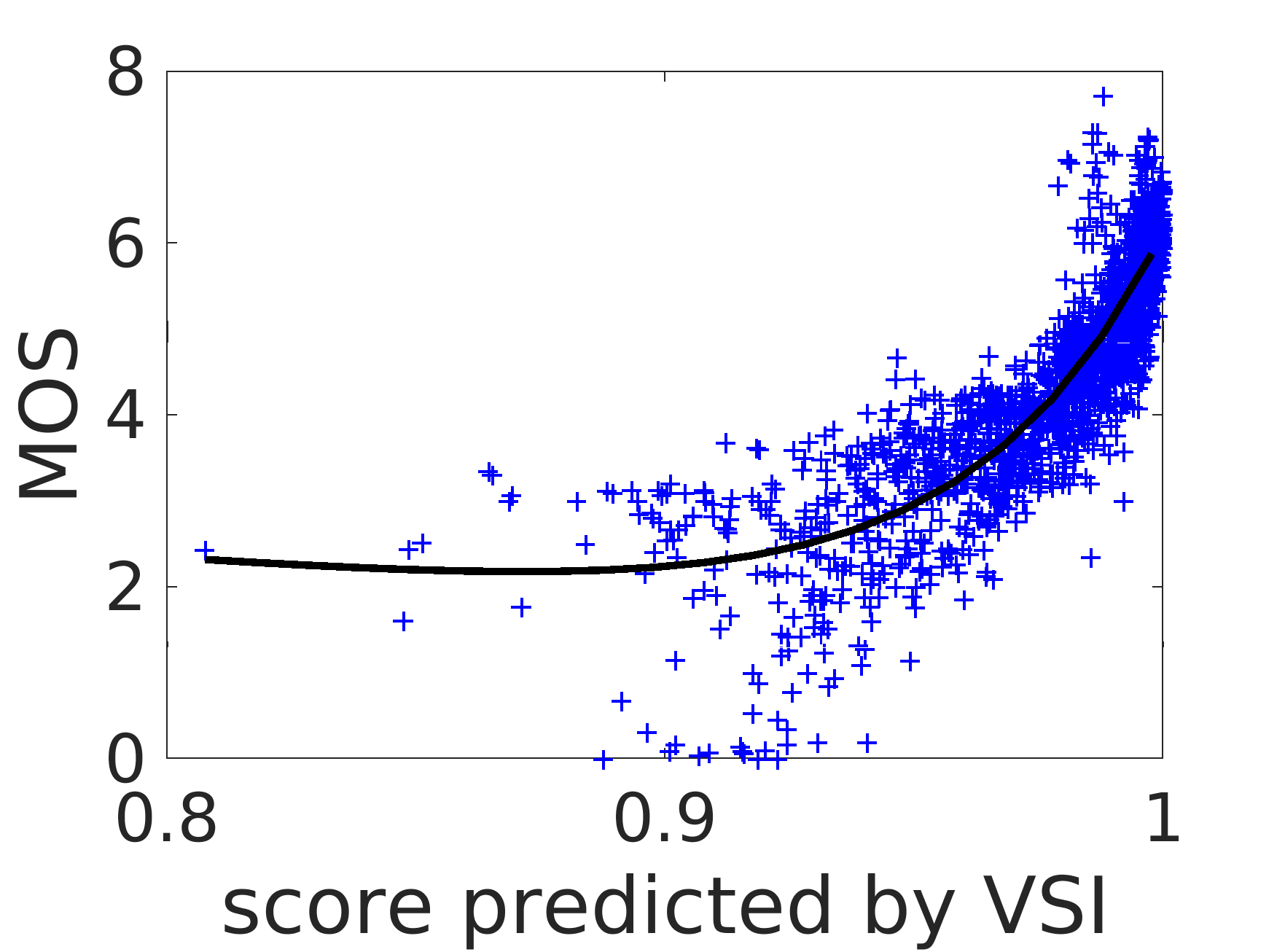}}
		&
		\subfloat[\label{fig:PridictMCSD}]{
			\includegraphics[width=.3\linewidth]{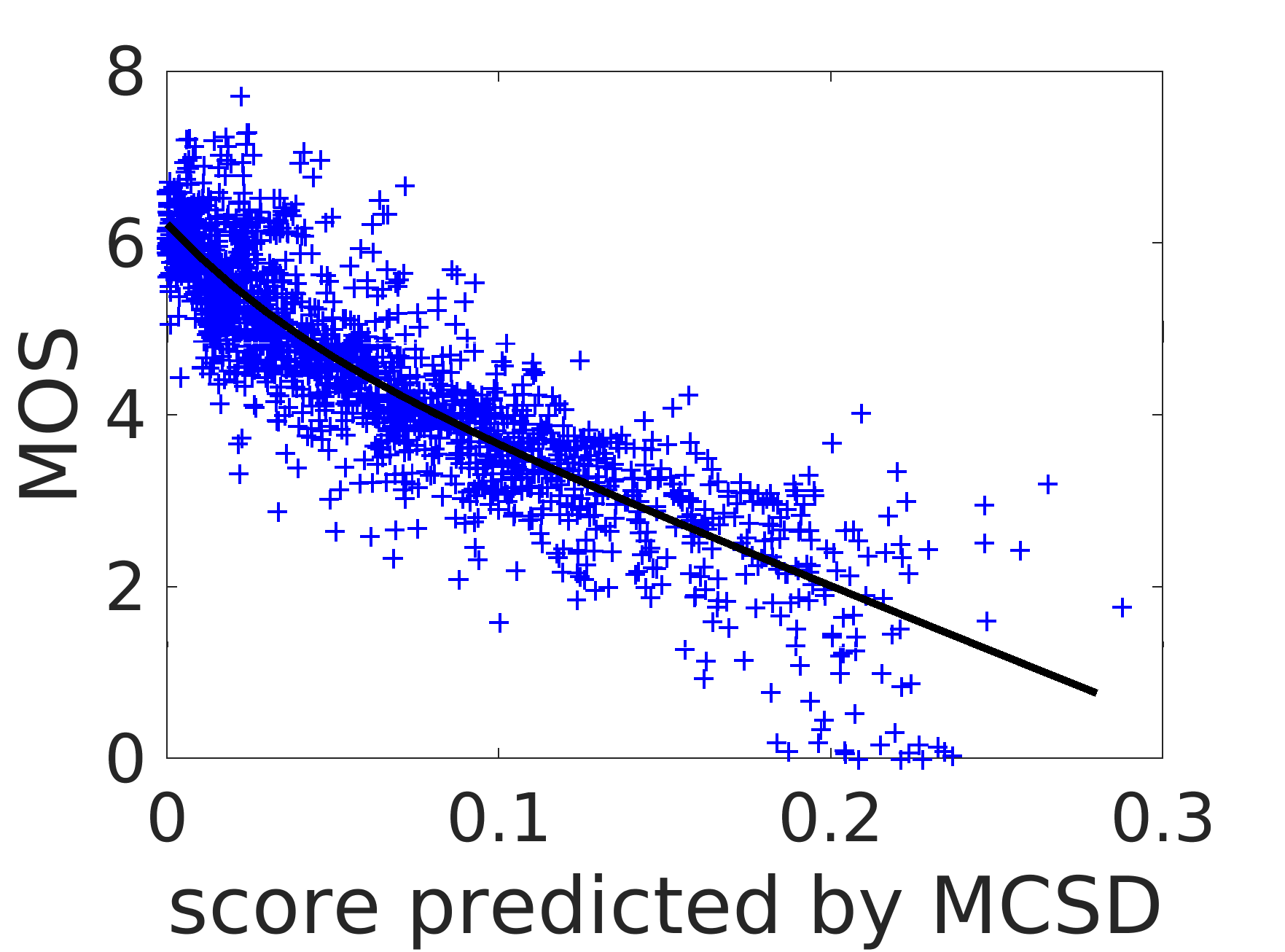}}
		&
		\subfloat[\label{fig:PridictVIF}]{
			\includegraphics[width=.3\linewidth]{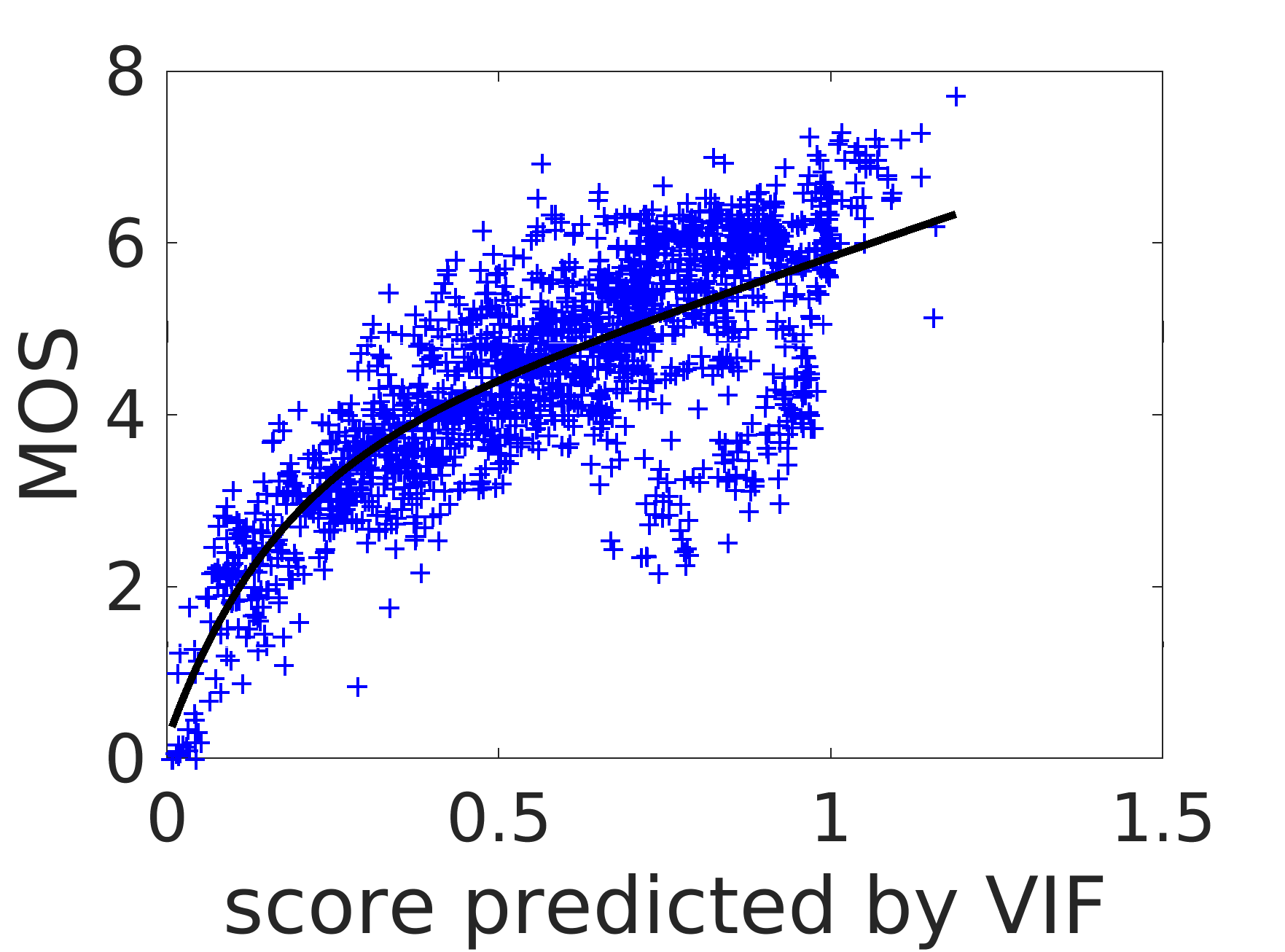}}\\

		\subfloat[\label{fig:PridictSRSIM}]{
			\includegraphics[width=.3\linewidth]{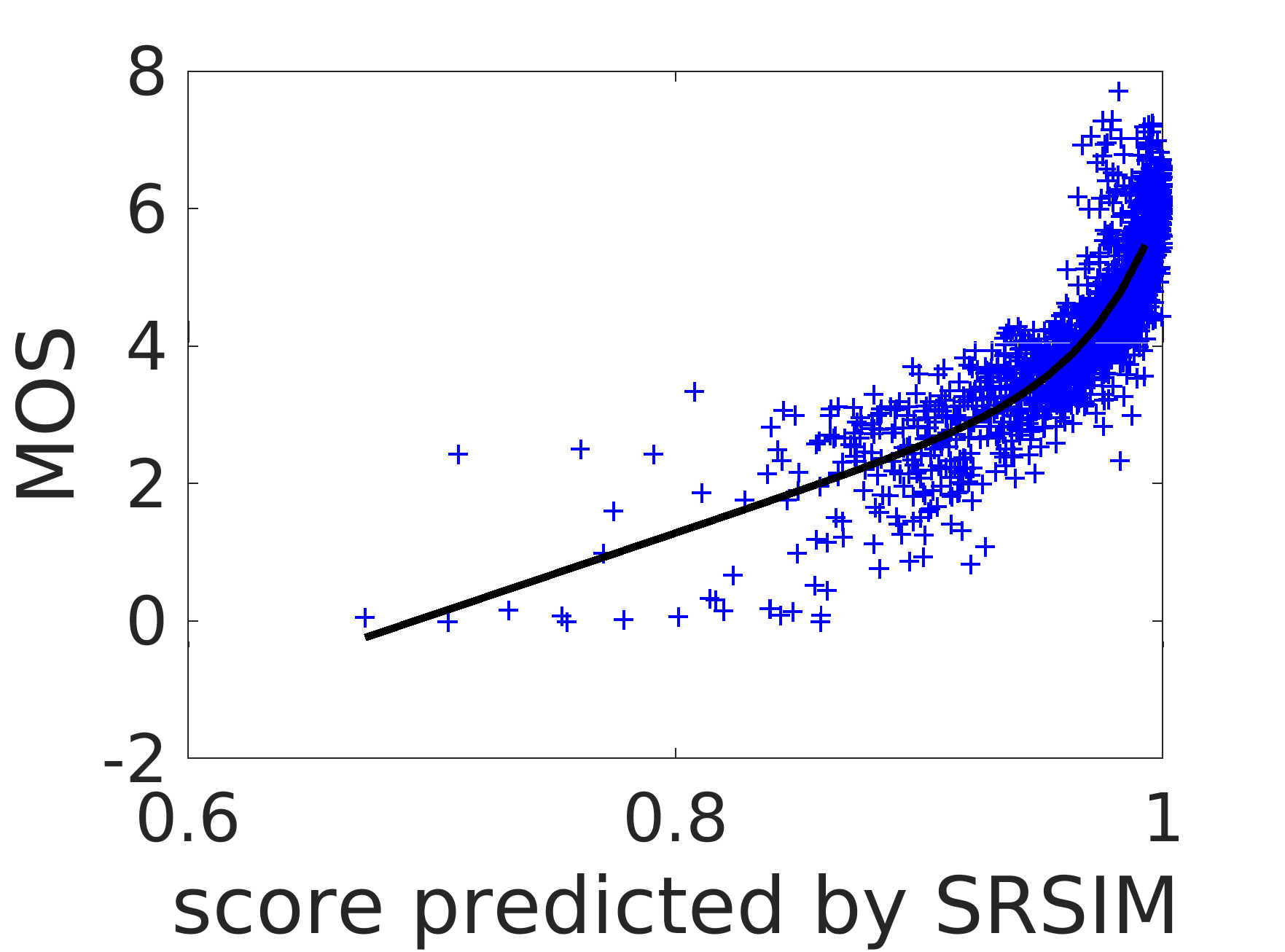}}
		&
		\subfloat[\label{fig:PridictVSP}]{
			\includegraphics[width=.3\linewidth]{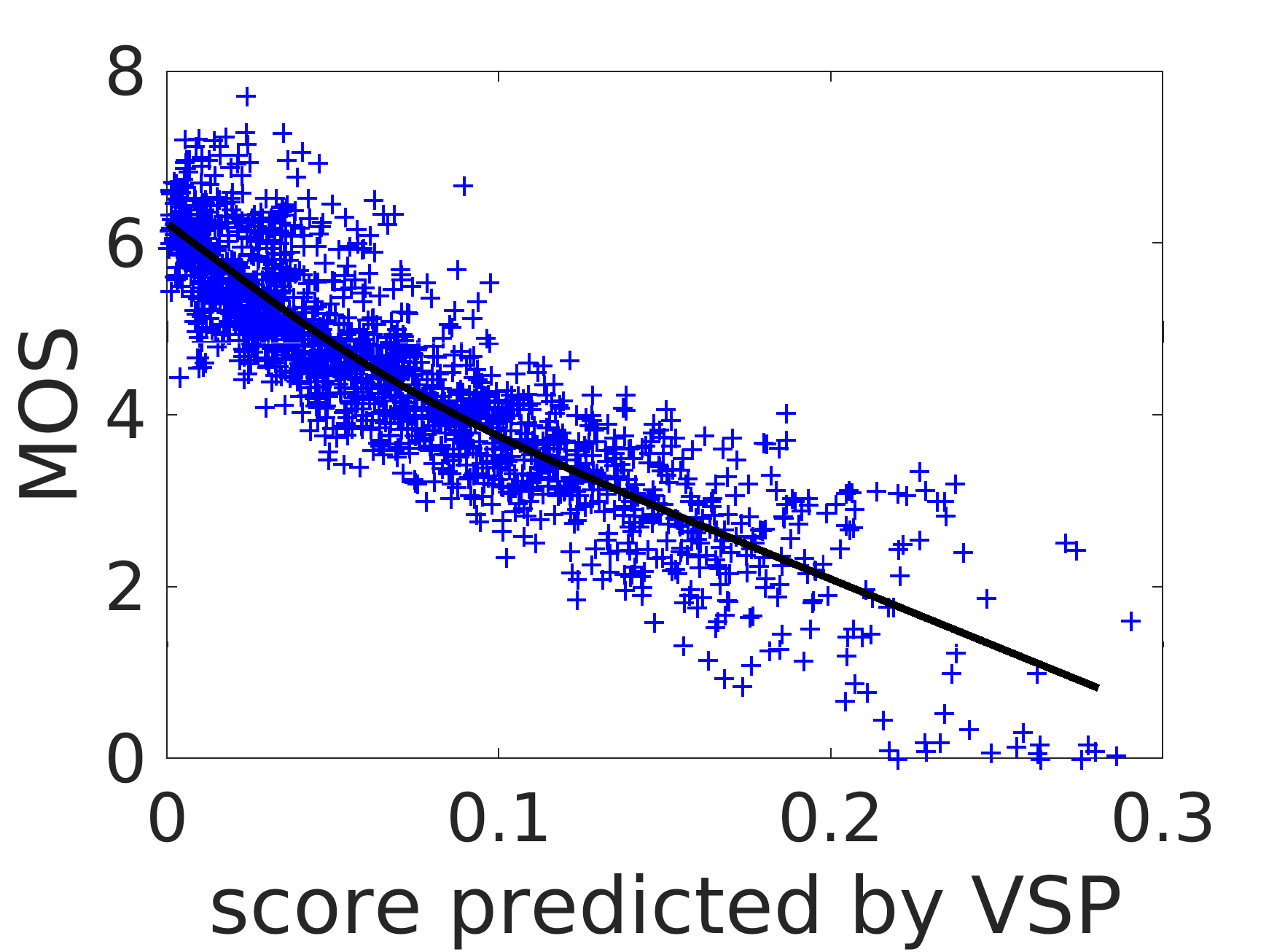}}
		&
		\subfloat[(Proposed)\label{fig:PridictCEQI}]{
			\includegraphics[width=.3\linewidth]{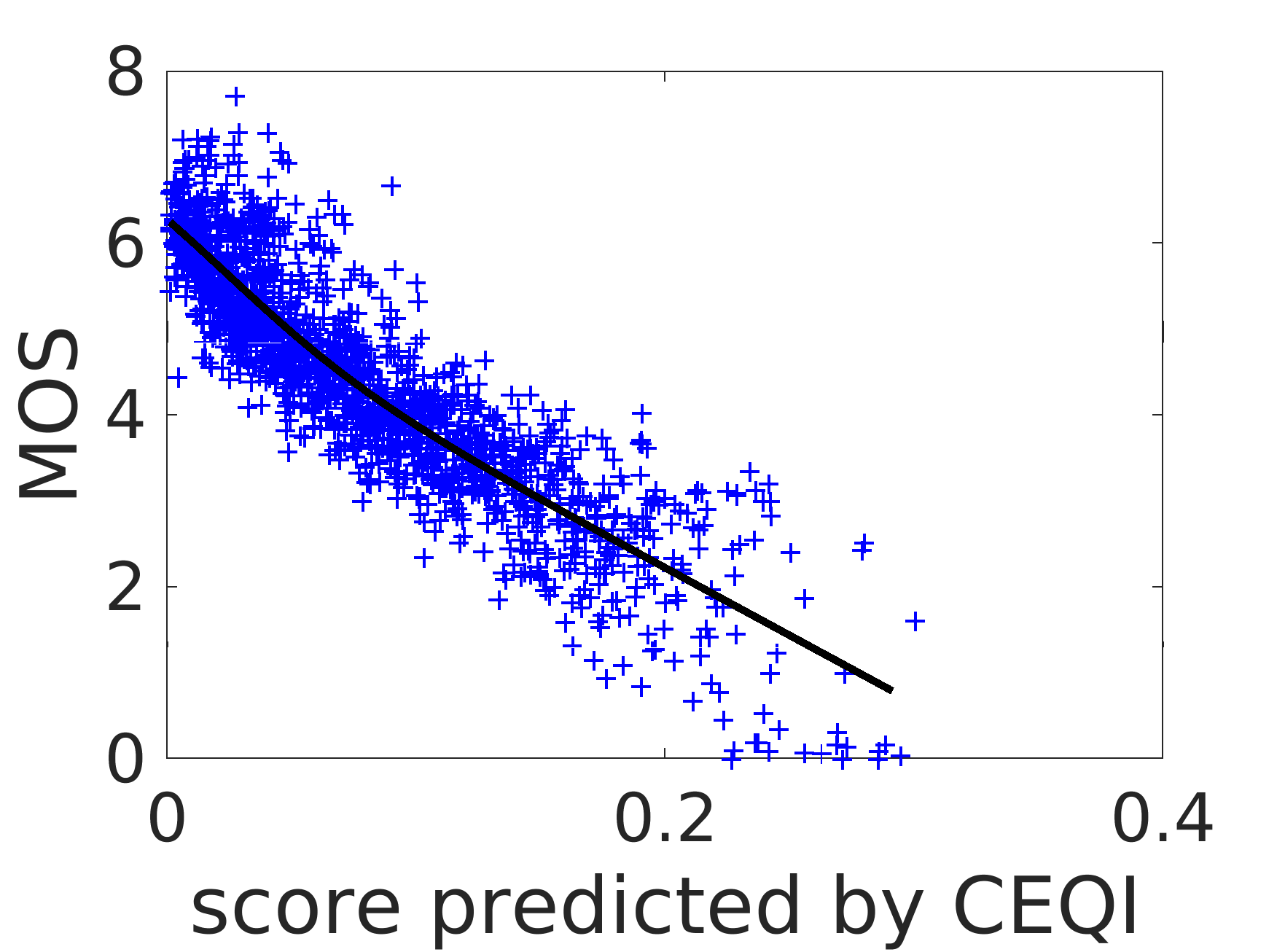}}\\
		
	\end{tabular}
	\caption{Scatter plots of the mean opinion scores (MOS/DMOS) versus scores predicted by different methods on the TID2008 database. The black curves are obtained by the nonlinear fitting based on Equation \ref{eq:DataFit} }
	\label{fig:ScatterPlot}
\end{figure}

Figure \ref{fig:ScatterPlot} shows scatter plots of the predicted scores for different IQA approaches with the MOS/DMOS values on the TID2008 database. These results show that CEQI's prediction is consistent compared to other methods, while providing a better correlation. We do not include PSNR because its predictions are very inconsistent. NQM is also discarded for the same reason, although its performance is not as inconsistent as PSNR.

Although the prime consideration of an IQA model is the performance of its prediction, having a low computational cost is also a desirable feature, especially for a real-time system. We evaluated the various IQA models with MATLAB R2017b using a computer equipped with an Intel(R) Core(TM) i5-4670 CPU with a 3.40GHz processor and 16GB of RAM. The MATLAB codes provided by the authors were used and elapsed time was recorded using the traditional \textit{tic-toc} function. The results of these tests are shown in Table \ref{Table:RunningTimeComparison}.  As expected, PSNR has the lowest computation time. Surprisingly, the gradient magnitude similarity division model can process 263.05 images per second with satisfactory performance (rank 4 as shown in Table \ref{Table:OverallRanking}). VIF shows very good performance for the LIVE database where it is the best-performing IQA, but it can only process 1.79 images per second on average, which makes it inappropriate for real-time systems or systems with low processing capability. On the other hand, CEQI takes 15.25 milliseconds to process an image, with the capability of processing 65.51 images per second. This frame rate meets the need for almost all kinds of real-time~systems. 

\begin{table}[H]
	\caption{Distortion-wise SROCC performance comparison of the IQA methods on three databases.}
	\label{Table:compareTableNoisewise}
	\vspace{-6pt}
	\centering
	%\resizebox{\columnwidth}{!}
	{	
		\includegraphics[width=.95\textwidth]{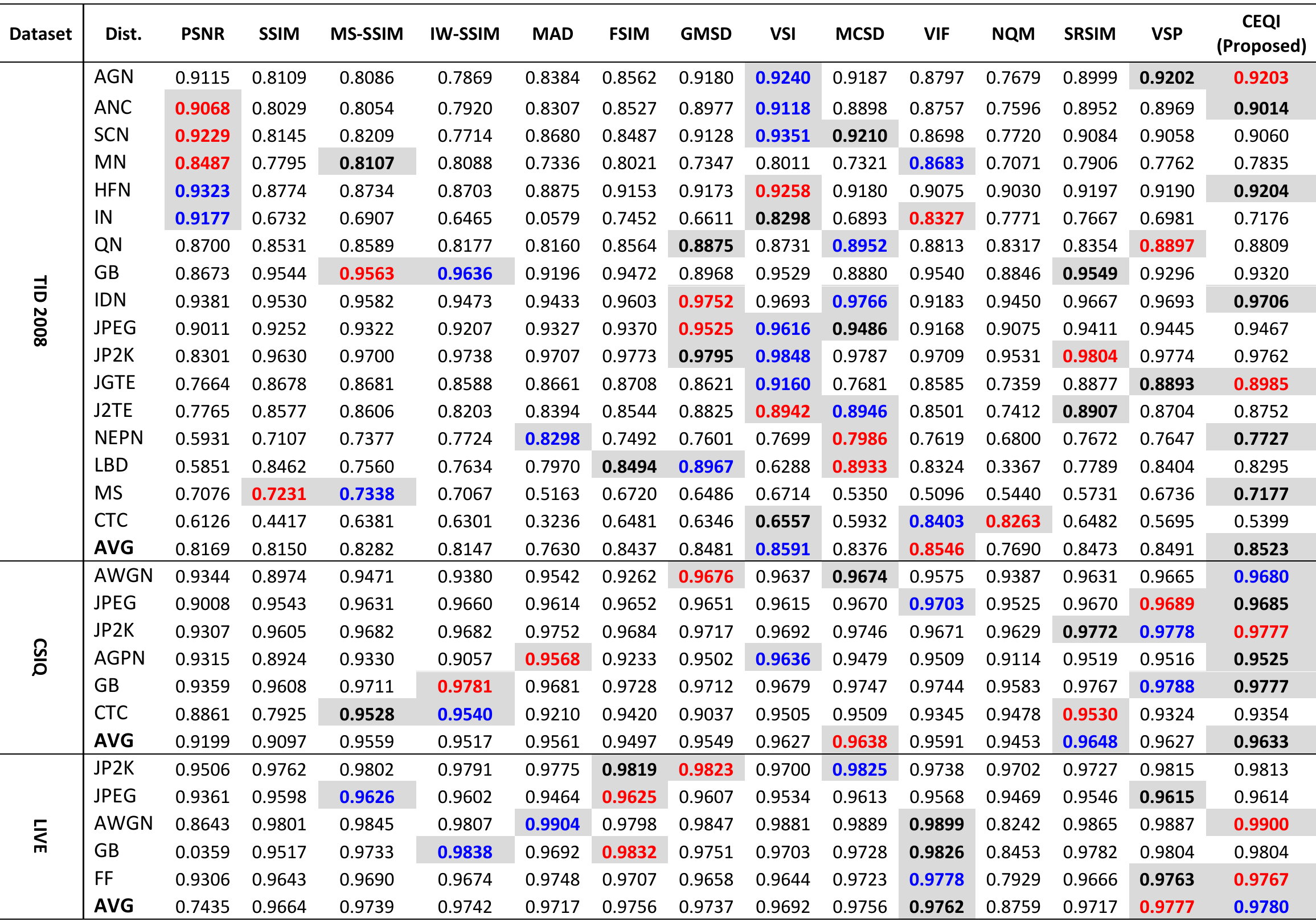}	
	}
	\begin{tabular}{@{}c@{}} 
		\multicolumn{1}{p{\linewidth-0.75cm}}{\footnotesize  -- For each row, the first, second and third-ranked performances are highlighted respectively in blue, red and black colors.}
		\\
		\multicolumn{1}{p{\linewidth-0.75cm}}{\footnotesize
			-- The acronyms for distortions are defined in Table \ref{Table:NoiseWiseDatasetDescription}, AVG refers to average performance over all noises in a database. }  
	\end{tabular}
\end{table}

\vspace{-12pt}

\begin{table}[H]
	\caption{Running time comparison of the IQA models.}
	\centering
	%\resizebox{\columnwidth}{!}
	{	
		\includegraphics[width=.80\textwidth]{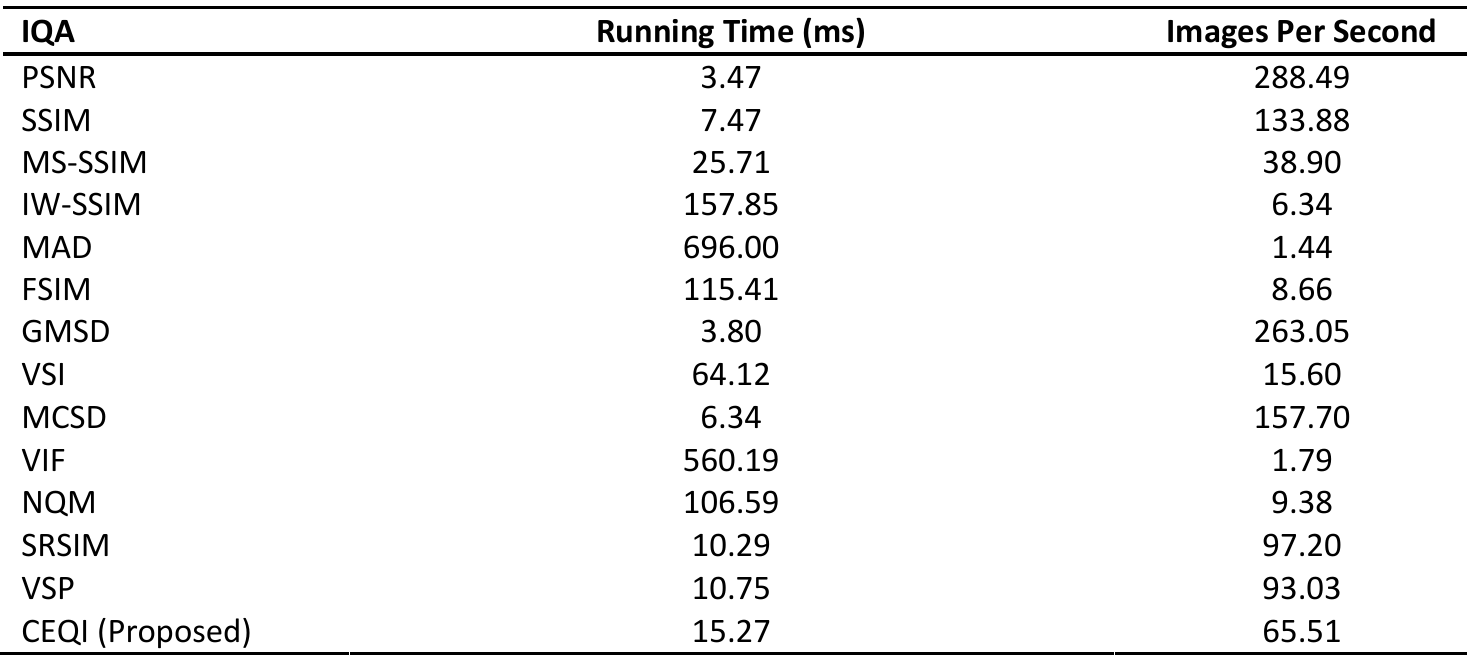}	
	}

	\label{Table:RunningTimeComparison}
\end{table}

%%%%%%%%%%%%%%%%%%%%%%%%%%%%%%%%%%%%%%%%%%%%%%%%%%%%%%%%%%%%%%%%%%%%%%%%%%%

%%%%%%%%%%%%%%%%%%%%%%%%%%%%%%%%%%%%%%%%%%%%%%%%%%%%%%%%%%%%%%%%%%%%%%%%%%%%%%%
\section{Conclusions}
\label{sec:con}
In this paper, we considered the center bias of HVS and proposed a full-reference image quality assessment method, CEQI, combining visual saliency and contrast. We placed extra emphasis on the center part of the image so that any degradation within the center region results in more effects than other regions. The proposed approach was compared with other state-of-the-art IQA models and it outperforms other competing methods in most cases. Comparing individual distortion types, the proposed method gives consistent scores. Additionally, the running time is suitable for real-time applications. The center emphasis makes the method more balanced and robust. We believe that this center emphasis will enhance the performance of other existing IQA models, including no-reference and reduced-reference approaches. In our future work, we will investigate these~possibilities.

%%%%%%%%%%%%%%%%%%%%%%%%%%%%%%%%%%%%%%%%%%
\section*{acknowledgments}
This work was supported by Institute for Information \& communications Technology Promotion(IITP) grant funded by the Korea government(MSIT) (No.2017-0-00294, Service mobility support distributed cloud technology)

%\nolinenumbers
%=====================================
\bibliographystyle{plainnat}

\bibliography{manuscript}

\end{document}